\documentclass{article}

\usepackage[preprint]{neurips_2024}

\makeatletter
\def\@noticestring{Preprint. \today.}
\makeatother

\usepackage[utf8]{inputenc}
\usepackage[T1]{fontenc}
\usepackage{hyperref}
\usepackage{url}
\usepackage{booktabs}
\usepackage{amsfonts}
\usepackage{amsmath}
\usepackage{nicefrac}
\usepackage{microtype}
\usepackage{graphicx}
\usepackage{placeins}
\usepackage{tikz}
\usetikzlibrary{arrows.meta, positioning, fit, backgrounds, calc}

\title{A Single-Layer Model Can Do Language Modeling\thanks{Code: \url{https://github.com/steve-z-wang/grounded-prediction-network}}}

\author{%
  Zanmin Wang \\
  \texttt{zanmin.wang@proton.me}
}

\begin{document}
\maketitle

\begin{abstract}
Modern language models scale depth by stacking layers, each holding
its own state --- a per-layer KV cache in transformers, a per-layer
matrix in Mamba, Gated DeltaNet (GDN), RWKV, and xLSTM. Biological
systems lean heavily on recurrence rather than on stacking. We ask
how far that shape can go on language modeling. We propose Grounded
Prediction Networks (GPN): one state vector revisited at every step
through a single recurrent block --- one FFN, one shared matrix
memory. At 130M parameters, a 1-layer GPN+M reaches FineWeb-Edu
perplexity $18.06$, within $13\%$ of a 12-layer Transformer++
($16.05$) and $18\%$ of a 10-layer GDN ($15.34$); a 2-layer variant
closes the gap to $6\%/11\%$. We do not match the deep
baselines. Because the working context is a single vector, we can
directly inspect its geometry: a persistent default-token direction,
a content-bearing horizon of tens of tokens, and memory heads that
split spontaneously into fast and slow retention pools.
\end{abstract}

\section{Introduction}

Sequence models scale depth by stacking layers, and that depth comes
with state. A transformer holds a per-layer KV cache; Mamba, GDN,
RWKV, and xLSTM hold a per-layer matrix state. Both computation and
memory are distributed across the stack: every layer keeps its own
copy of ``what the model knows so far.''

Biological networks lean heavily on recurrence rather than on
stacking, with depth coming from revisiting the same circuitry
across time. We ask whether the same shape is viable for language
modeling: how far can a single shallow recurrent layer go?

A Grounded Prediction Network (GPN) is our answer. One state vector,
one FFN, one shared matrix memory --- revisited at every step. Each
step grounds the previous state with the observed token, predicts
the next state via the FFN and a read from the matrix memory, and
decodes the next-token distribution. There is no stack; depth comes
from time. We call this loop \emph{grounded prediction} and unpack it
in \S\ref{sec:gp}.

At 130M parameters, a 1-layer GPN+M reaches FineWeb-Edu perplexity
$18.06$, within $13\%/18\%$ of a 12-layer Transformer++ ($16.05$)
and a 10-layer GDN ($15.34$); a 2-layer variant closes the gap to
$6\%/11\%$. The shallow model does not match the deep baselines. And
because the entire state is one vector, we can observe what the
trained model puts there: a persistent default-token direction, a
content-bearing horizon of tens of tokens, and memory heads that
split into fast and slow retention pools.

\paragraph{Contributions.}
\begin{enumerate}
\setlength{\itemsep}{0pt}
\item \textbf{Architecture.} GPN/GPN+M --- one state vector, one
FFN, one centralized matrix memory, revisited at every step rather
than stacked (\S\ref{sec:architecture}).
\item \textbf{A conceptual view.} The state's prediction of the
next step can be viewed as part of the encoding of that step ---
a single vector plays both roles, with grounding completing the
encoding (\S\ref{sec:gp}).
\item \textbf{Evidence.} At 130M parameters, a 1-layer GPN+M is
within $13\%/18\%$ FineWeb-Edu PPL of a 12-layer Transformer++ and
a 10-layer GDN; a 2-layer variant closes the gap to $6\%/11\%$
(\S\ref{sec:results}).
\item \textbf{Observability.} A persistent default-token direction,
a tens-of-token content horizon, and a spontaneous fast/slow memory
split among architecturally-identical heads --- structures a layered
stack would have to dig out of residual streams
(\S\ref{sec:analysis}).
\end{enumerate}

\section{Related Work}

\paragraph{Recurrent and linear-recurrent LMs.} RNNs
\cite{hochreiter1997long}, SSMs \cite{gu2021efficiently,gu2023mamba},
and gated linear recurrences
\cite{yang2023gatedlinear,dao2024transformers,yang2024gated} maintain
a per-layer state that evolves in time, usually a matrix that plays
both working-context and associative-memory roles in one object.
Griffin/Hawk \cite{de2024griffin}, RWKV-7 \cite{peng2025rwkv7}, and
xLSTM \cite{beck2024xlstm} are recent entries in the same line. GPN
differs in architectural shape: one wide shallow layer instead of a
stack, and a single centralized associative memory shared across
layers instead of per-layer recurrent state.

\paragraph{Associative memory.} Linear attention
\cite{katharopoulos2020transformers} and the delta rule
\cite{schlag2021linear,yang2024parallelizing,yang2024gated} view
attention as fast-weight associative storage. GPN+M uses the gated
delta rule as a single shared memory rather than a per-layer KV
cache or per-layer SSM state.

\paragraph{Short-term / long-term memory splits.} Several recent
architectures separate short-term and long-term memory by design:
Infini-attention \cite{munkhdalai2024infini}, Hymba
\cite{dong2024hymba}, Titans \cite{behrouz2025titans}, and
nested/continuum-memory learning \cite{behrouz2025nested} all pair
an attention-based local buffer with a separate compressive or
recurrent long-term store. GPN+M does not architect such a split:
it has one persistent recurrent state and one centralized associative
memory, with no attention window at either side. The fast/slow split
among its memory heads emerges in training rather than being built
in (\S\ref{sec:horizon}).

\section{Grounded Prediction}
\label{sec:gp}

\paragraph{Notation.} Let $x_1, \ldots, x_T$ be a token stream. A
step of GPN holds the same working-context vector at two points:
\begin{itemize}
\setlength{\itemsep}{0pt}
\item $s^p_t \in \mathbb{R}^d$ --- the \emph{predicted state}. One
vector, produced at the end of step $t$ and carried into step
$t+1$ as the model's forecast for the next state, decoded
downstream into a distribution over $x_{t+1}$.
\item $s^g_t \in \mathbb{R}^d$ --- the \emph{grounded state}.
Intermediate: Ground$(s^p_{t-1}, x_t)$ produces it; Predict
consumes it.
\end{itemize}
We write $\bar s = \mathbb{E}_t[s^p_t]$ for the mean of the
persistent state, and use $s$ without superscripts when the stage
is clear. Structurally $s$ is a residual flow: Ground and Predict
each add their contribution through gated residuals
(\S\ref{sec:architecture}). This is analogous to a transformer's
residual stream, except the additions accumulate across \emph{time}
(one step per token) rather than across \emph{depth} (one layer at
a time).

\begin{figure}[!ht]
\centering
\begin{tikzpicture}[
  font=\small,
  state/.style={circle, draw, thick, minimum size=8mm, inner sep=0pt, fill=white},
  op/.style={rectangle, draw, rounded corners, thick, minimum height=7mm, minimum width=11mm, inner sep=2pt, fill=white},
  tok/.style={font=\small\itshape},
  >={Stealth[length=2mm]},
  node distance=6mm and 6mm,
  thick
]
  \node[state] (s0) {$s^p_{t-1}$};
  \node[op, right=of s0] (g1) {Ground};
  \node[state, right=of g1] (gr1) {$s^g_t$};
  \node[op, right=of gr1] (p1) {Predict};
  \node[state, right=of p1] (s1) {$s^p_t$};
  \node[op, right=of s1] (g2) {Ground};
  \node[state, right=of g2] (gr2) {$s^g_{t+1}$};
  \node[op, right=of gr2] (p2) {Predict};
  \node[state, right=of p2] (s2) {$s^p_{t+1}$};

  \draw[->] (s0) -- (g1);
  \draw[->] (g1) -- (gr1);
  \draw[->] (gr1) -- (p1);
  \draw[->] (p1) -- (s1);
  \draw[->] (s1) -- (g2);
  \draw[->] (g2) -- (gr2);
  \draw[->] (gr2) -- (p2);
  \draw[->] (p2) -- (s2);

  \node[tok, above=5mm of g1] (x1) {$x_t$};
  \node[tok, above=5mm of g2] (x2) {$x_{t+1}$};
  \draw[->] (x1) -- (g1);
  \draw[->] (x2) -- (g2);

  \node[op, below=5mm of s1] (d1) {Decode};
  \draw[->] (s1) -- (d1);
  \node[below=3mm of d1, font=\small] (l1) {$\mathcal{L}_t$};
  \draw[->] (d1) -- (l1);

  \node[op, below=5mm of s2] (d2) {Decode};
  \draw[->] (s2) -- (d2);
  \node[below=3mm of d2, font=\small] (l2) {$\mathcal{L}_{t+1}$};
  \draw[->] (d2) -- (l2);
\end{tikzpicture}
\caption{The grounded-prediction loop, unrolled for two steps. The
predicted state $s^p_{t-1}$ is grounded by the observed token $x_t$
to produce the grounded state $s^g_t$; Predict turns $s^g_t$ into
$s^p_t$, the forecast for $x_{t+1}$, which is decoded for the loss
and carried forward as input to the next Ground. Both $s^p$ and
$s^g$ are the same working-context vector at different points in
the step. Concrete realizations of Ground, Predict, and Decode ---
with and without the matrix memory --- appear in
\S\ref{sec:architecture} and Fig.~\ref{fig:arch}.}
\label{fig:loop}
\end{figure}
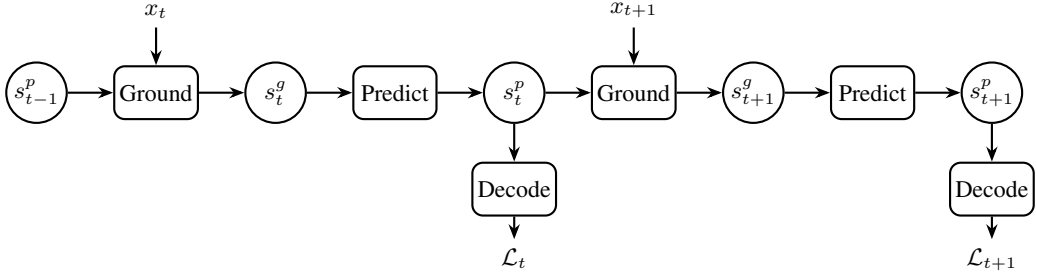

\paragraph{The loop.} Each step grounds the previous state with the
observed token, predicts the next state, and decodes:
\begin{align}
s^g_t &= \text{Ground}(s^p_{t-1}, x_t) \label{eq:ground}\\
s^p_t &= \text{Predict}(s^g_t) \label{eq:predict}\\
\mathcal{L}_t &= \text{Loss}\big(\text{Decode}(s^p_t),\; x_{t+1}\big). \label{eq:loss}
\end{align}
Training minimizes $\sum_t \mathcal{L}_t$; no auxiliary objective is
used. For discrete tokens, Decode is a linear head and Loss is
cross-entropy. \S\ref{sec:architecture} gives concrete realizations
of Ground, Predict, and Decode, first without a matrix memory (GPN)
and then with one (GPN+M).

\paragraph{Prediction as part of encoding.} The state $s^p_t$ can
be viewed in two ways. As a \emph{forecast}, it is the model's
prediction for the next state, decoded downstream into a
distribution over $x_{t+1}$. As an \emph{encoding}, it summarizes
$x_1, \ldots, x_t$ --- the model's record of what has been observed
so far, awaiting the next observation. The two views are not two
separate state-like objects; they are the same vector at different
moments in the loop. The prediction leaving step $t$ is the partial
encoding entering step $t+1$, which the new token grounds and
completes. Once grounded, the encoding is projected forward again
as the next prediction --- and the loop continues.

\FloatBarrier

\section{Architecture}
\label{sec:architecture}

The state $s^p_t \in \mathbb{R}^d$ is a single recurrent vector with
$d$ on the order of thousands; for our 130M GPN+M, $d = 2496$. We
define GPN first as the bare primitive with just an FFN-based
Predict, then extend it to GPN+M by adding the matrix memory.

\subsection{GPN: the bare primitive (no memory)}

\paragraph{Ground.} The token $x_t$ is embedded and fused into
$s^p_{t-1}$ through a gated residual:
\begin{align}
s^g_t &= \text{Gate}_f(s^p_{t-1}) \odot s^p_{t-1}
   + W_{\text{fuse}} \cdot \text{emb}(x_t),
\label{eq:ground-gpn}
\end{align}
where $\text{Gate}_f(\cdot) = \sigma\!\left(W_f \cdot \text{RMSNorm}(\cdot)\right)$
is a learnable fuse gate ($\sigma$ is the elementwise sigmoid).

\paragraph{Predict.} An FFN refines the grounded state, again with a
gated residual:
\begin{align}
s^p_t &= \text{Gate}_p(s^g_t) \odot s^g_t
   + \text{FFN}(s^g_t),
\label{eq:predict-gpn}
\end{align}
where $\text{FFN}(x) = (\text{SiLU}(W_1^g x) \odot W_1^v x)\, W_2$ is
a SwiGLU FFN with RMSNorm pre-norm, and $\text{Gate}_p$ is a
learnable predict gate of the same form as $\text{Gate}_f$.

\paragraph{Decode.} The decoder is a single linear head:
\begin{align}
\mathcal{L}_t &= \text{CE}\big(W_{\text{dec}}\,
   \text{RMSNorm}(s^p_t),\; x_{t+1}\big).
\label{eq:decode-gpn}
\end{align}

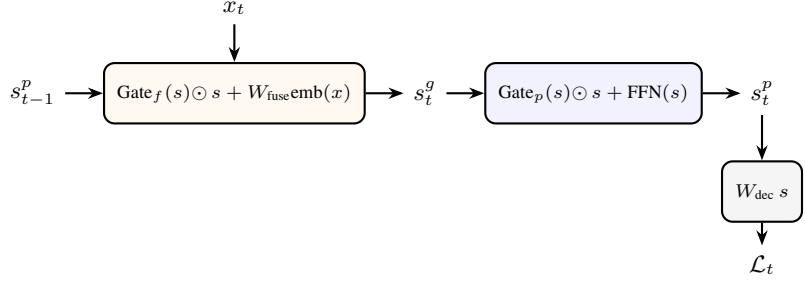
\begin{figure}[!ht]
\centering
\begin{tikzpicture}[
  font=\small,
  block/.style={rectangle, draw, rounded corners, thick, minimum height=8mm, inner sep=5pt, align=center, fill=blue!5, font=\scriptsize},
  gblock/.style={rectangle, draw, rounded corners, thick, minimum height=8mm, inner sep=5pt, align=center, fill=orange!5, font=\scriptsize},
  dblock/.style={rectangle, draw, rounded corners, thick, minimum height=8mm, inner sep=5pt, align=center, fill=gray!8, font=\scriptsize},
  sstate/.style={font=\small},
  tok/.style={font=\small\itshape},
  lbl/.style={font=\scriptsize, inner sep=1pt, fill=white},
  >={Stealth[length=2mm]},
  thick
]
  \node[sstate] (sprev) {$s^p_{t-1}$};
  \node[gblock, right=5mm of sprev]
    (g) {$\text{Gate}_f(s){\odot}\,s + W_{\text{fuse}}\text{emb}(x)$};
  \node[sstate, right=5mm of g] (gr) {$s^g_t$};
  \node[block, right=5mm of gr]
    (p) {$\text{Gate}_p(s){\odot}\,s + \text{FFN}(s)$};
  \node[sstate, right=5mm of p] (spost) {$s^p_t$};

  \draw[->] (sprev) -- (g);
  \draw[->] (g) -- (gr);
  \draw[->] (gr) -- (p);
  \draw[->] (p) -- (spost);

  \node[tok, above=5mm of g] (x) {$x_t$};
  \draw[->] (x) -- (g);

  \node[dblock, below=6mm of spost] (d) {$W_{\text{dec}}\,s$};
  \node[below=3mm of d, font=\small] (l) {$\mathcal{L}_t$};
  \draw[->] (spost) -- (d);
  \draw[->] (d) -- (l);
\end{tikzpicture}
\caption{One step of GPN (no memory), horizontal. Blocks show the
operation formulas directly; inside a block, $s$ and $x$ refer to
the block's input (the state flowing in). $s^p_{t-1}$ enters from
the left; $s^p_t$ exits on the right and becomes the input to the
next step. Decode reads $s^p_t$ downward for the loss
$\mathcal{L}_t = \text{CE}(\cdot, x_{t+1})$. Implementation details
(RMSNorm pre-norm before the decode head, RMSNorm inside Gate and
FFN) are omitted for brevity; see \S\ref{sec:architecture}.}
\label{fig:arch-gpn}
\end{figure}

\FloatBarrier

\subsection{GPN+M: adding a matrix memory}

GPN+M keeps Ground and Decode unchanged and extends Predict with a
matrix-memory read:
\begin{align}
s^p_t &= \text{Gate}_p(s^g_t) \odot s^g_t
   + \text{FFN}(s^g_t)
   + \text{MemRead}(s^g_t, M_t).
\label{eq:predict-gpnm}
\end{align}
The matrix memory $M \in \mathbb{R}^{H \times d_k \times d_v}$ with
$H$ heads is updated by the gated delta rule:
\begin{align}
k_t &= \ell_2\text{-norm}(\text{ReLU}(W_k\, s^g_{t-1})),\qquad
v_t = W_v\, s^g_t
\label{eq:kv}\\
\beta_t &= \sigma(W_\beta\, s^g_t),\qquad
\gamma_t = -\exp(A)\cdot\text{softplus}(W_a\, s^g_t)
\label{eq:beta-gamma}\\
M_t &= e^{\gamma_t} \odot M_{t-1}
   + \beta_t\, \bigl(k_t \otimes (v_t - M_{t-1} k_t)\bigr).
\label{eq:mwrite}
\end{align}
The key comes from the previous grounded state $s^g_{t-1}$ and the
value from the current grounded state $s^g_t$, so the rule explicitly
learns $s^g_{t-1} \to s^g_t$ transitions. Memory reads run in parallel
with the FFN: the query $q_t =
\ell_2\text{-norm}(\text{ReLU}(W_q\, s^g_t))$ retrieves
\begin{align}
\text{MemRead}(s^g_t, M_t) \;=\;
   \text{silu}(W_\text{rg}\, s^g_t)
   \odot \text{RMS}(M_t\, q_t)\, W_o,
\label{eq:mread}
\end{align}
which is added into the state update alongside the FFN output
(Eq.~\ref{eq:predict-gpnm}).

\begin{figure}[!ht]
\centering
\begin{tikzpicture}[
  font=\small,
  block/.style={rectangle, draw, rounded corners, thick, minimum height=10mm, inner sep=5pt, align=center, fill=blue!5, font=\scriptsize},
  gblock/.style={rectangle, draw, rounded corners, thick, minimum height=10mm, inner sep=5pt, align=center, fill=orange!5, font=\scriptsize},
  dblock/.style={rectangle, draw, rounded corners, thick, minimum height=8mm, inner sep=5pt, align=center, fill=gray!8, font=\scriptsize},
  pstate/.style={font=\small, align=center},
  tok/.style={font=\small\itshape},
  >={Stealth[length=2mm]},
  thick
]
  \node[pstate] (sprev) {$s^p_{t-1}$,\\[-1pt]$M_{t-1}$};
  \node[gblock, right=5mm of sprev]
    (g) {$\text{Gate}_f(s){\odot}\,s + W_{\text{fuse}}\text{emb}(x)$};
  \node[pstate, right=5mm of g] (gr) {$s^g_t$,\\[-1pt]$M_{t-1}$};
  \node[block, right=5mm of gr]
    (p) {$M \leftarrow \text{MemWrite}(s, M)$\\
         $\text{Gate}_p(s){\odot}\,s + \text{FFN}(s) + \text{MemRead}(s, M)$};
  \node[pstate, right=5mm of p] (spost) {$s^p_t$,\\[-1pt]$M_t$};

  \draw[->] (sprev) -- (g);
  \draw[->] (g) -- (gr);
  \draw[->] (gr) -- (p);
  \draw[->] (p) -- (spost);

  \node[tok, above=5mm of g] (x) {$x_t$};
  \draw[->] (x) -- (g);

  \node[dblock, below=6mm of spost] (d) {$W_{\text{dec}}\,s$};
  \node[below=3mm of d, font=\small] (l) {$\mathcal{L}_t$};
  \draw[->] (spost) -- (d);
  \draw[->] (d) -- (l);
\end{tikzpicture}
\caption{One step of GPN+M, horizontal. Each state bundle carries
both the working context $s^p$ and the matrix memory $M$; inside a
block, $s, M$ refer to the block's input. Ground updates only
$s^p$. Predict (blue, two lines) first updates $M$ via MemWrite
(gated delta rule) and then computes the new state as gated
residual $+$ FFN $+$ MemRead. The bundle $(s^p_t, M_t)$ exits on
the right and becomes the input to the next step. Decode reads
$s^p_t$ downward for the loss. GPN (Fig.~\ref{fig:arch-gpn}) is
this diagram with $M$ and MemWrite/MemRead removed.}
\label{fig:arch}
\end{figure}
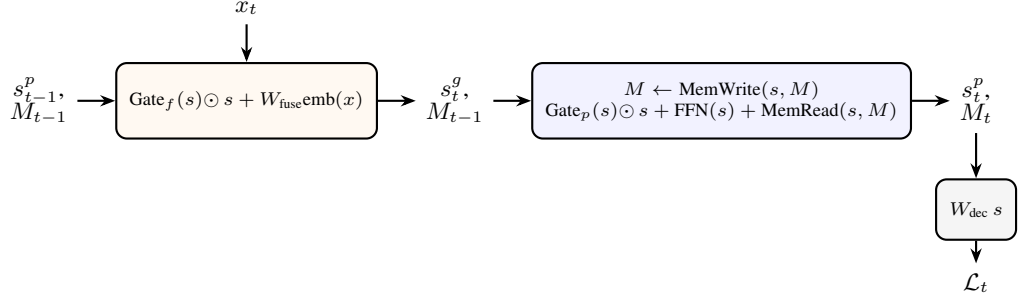

\paragraph{Predict = FFN $+$ memory, two centralized predictors.}
Predict is the sum of two predictor functions. The FFN predicts
from parameters trained offline; the memory read predicts from
parameters written online --- whatever keys and values the last few
hundred steps have put into $M$. Both enter the state update
additively, as peers (Fig.~\ref{fig:arch}). Both are also
\emph{centralized}: there is one $M$ shared across all layers,
written once per token and read by every layer.

\paragraph{Configurations.} We train GPN+M at 130M parameters in
two depths: a 1-layer variant ($d=2496$, $d_{\text{ffn}}=6656$, $H=15$
memory heads with $d_k=128,\ d_v=256$) --- the headline single-layer
configuration, also used for the analyses in \S\ref{sec:analysis} ---
and a 2-layer variant ($d=1920$) that trades the simplest
single-layer story for a smaller PPL gap. Vocabulary is $32$K. GPN
is the same architecture with MemRead removed. Baselines are a
12-layer Transformer++ (RoPE, RMSNorm, SwiGLU, no bias) and a
10-layer Gated DeltaNet, both at matched 130M parameters.

\FloatBarrier

\section{Training}

Models are trained on FineWeb-Edu-10B tokenized with the LLaMA-2 tokenizer
(vocab 32k), sequence length 2048, batch 128 at 10k optimization steps.
Optimizer AdamW with peak LR $6 \cdot 10^{-4}$, cosine schedule to 10\% of
peak after 100-step warmup, weight decay 0.1. Transformer and GDN run in parallel along the sequence. GPN/GPN+M
run sequentially in Python across time; for 1-layer configurations
we use full BPTT across the 2048-token window, for 2-layer we
truncate BPTT at length 512 to fit memory. All models use bfloat16
mixed precision.

\FloatBarrier
\section{Results}
\label{sec:results}

\begin{table}[!ht]
\caption{Held-out perplexity at 130M parameters. \textbf{FineWeb}:
last (unused) bin of shuffled FineWeb-Edu-10B, all positions of a
2048-token window. \textbf{LAMBADA}: EleutherAI split (5153
examples), tokenized with Llama-2; accuracy is top-1 match of the
first subword of the last word given its prefix, PPL is $\exp$(mean
CE over the last word's subwords). ``Mem.\ cells'' counts cells in
the architecture's associative/recurrent memory.}
\label{tab:main}
\centering
\begin{tabular}{lrrrrrr}
\toprule
& & & & FineWeb & \multicolumn{2}{c}{LAMBADA} \\
\cmidrule(lr){5-5}\cmidrule(lr){6-7}
Model & Layers & State dim & Mem. cells & PPL $\downarrow$ & Acc $\uparrow$ & PPL $\downarrow$ \\
\midrule
Transformer++ & 12 & 768 & --- (KV) & 16.05 & \textbf{0.223} & 18.99 \\
GDN & 10 & 768 & 1.97M (distributed) & \textbf{15.34} & 0.208 & \textbf{17.23} \\
\midrule
GPN+M (ours) & 1 & 2496 & 492K (centralized) & 18.06 & 0.146 & 33.45 \\
GPN+M (ours) & 2 & 1920 & 360K (centralized) & 16.95 & 0.160 & 27.39 \\
GPN (no memory) & 1 & 3184 & 0 & 23.51 & 0.015 & 263.6 \\
\bottomrule
\end{tabular}
\end{table}

We train one seed per model with no scaling sweeps; these numbers
establish viability, not state-of-the-art. 1L GPN+M reaches FineWeb
PPL $18.06$, $13\%$ above a 12-layer Transformer++ ($16.05$) and
$18\%$ above a 10-layer GDN ($15.34$); a 2-layer variant brings the
gap to $6\%/11\%$ (PPL $16.95$).

\section{Analysis}
\label{sec:analysis}

Because the state is one vector, we measure it directly. All
analyses run on the trained 1L GPN+M checkpoint (130M) using
held-out data, with no additional training. Three structures appear
that were not asked for by the architecture: a high-rank,
near-decorrelated state geometry that uses its full width without
any decorrelation regularizer (\S\ref{sec:geometry}); a single
persistent direction inside the state that decodes to the training
unigram (\S\ref{sec:dc}); and a fast/slow retention split among
architecturally-identical memory heads (\S\ref{sec:horizon}).

\subsection{Geometry of the state}
\label{sec:geometry}

The state is wide: $d = 2496$, versus $d = 768$ for the
transformer and GDN baselines at matched parameter count. Most of
the parameter budget sits in state width rather than depth. Mean
magnitude $\|s^p_t\| \approx 132$, tightly concentrated across the
sequence ($\sigma_{\|s\|} \approx 9$).

\begin{figure}[!ht]
\centering
\includegraphics[width=\linewidth]{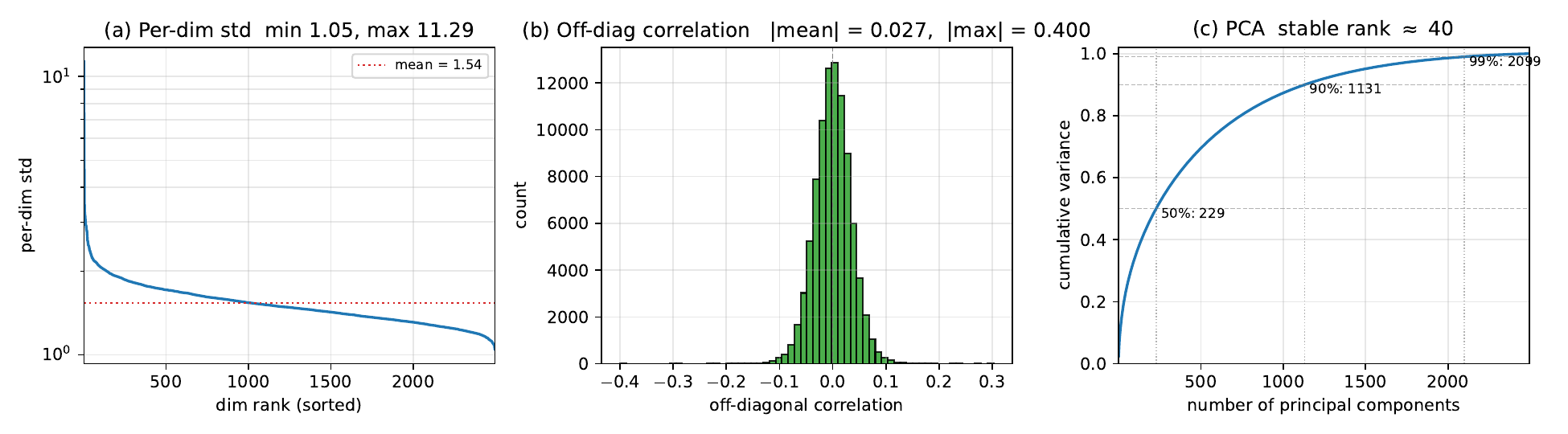}
\caption{State geometry (1L GPN+M, 130M).
\textbf{(a)}~Sorted per-dim std: heavy tail, no dead dimensions
(min $1.05$, max $11.3$).
\textbf{(b)}~Off-diagonal correlations (300-dim sample):
$|\text{mean}|=0.03$.
\textbf{(c)}~Cumulative variance: $1131$ PCs for $90\%$, $2099$
for $99\%$; stable rank $\approx 40$. High-rank, near-decorrelated,
no dead dimensions --- with no explicit regularizer.}
\label{fig:state_geometry}
\end{figure}

Three structural properties appear without any explicit regularizer
(Fig.~\ref{fig:state_geometry}). \textbf{(a)} Per-dimension std
ranges from $1.05$ to $11.3$, zero dead dimensions, heavy tail.
\textbf{(b)} Mean absolute off-diagonal correlation is $0.027$, max
$0.61$ --- dimensions are near-independent. \textbf{(c)} PCA on 16k
state vectors needs 1131 of 2496 components for $90\%$ variance,
2099 for $99\%$. Stable rank is 40, participation ratio 398: the
content signal is not collapsed onto a small subspace.

\paragraph{A dominant mean direction.} The state's mean
$\bar s = \mathbb{E}_t[s^p_t]$ has magnitude $\|\bar s\| \approx 107$,
$58\%$ of the typical state magnitude. The state is not a
zero-centered fluctuation but a large fixed offset plus a smaller
fluctuation around it. $\bar s$ is a \emph{global} property of the
trained model: per-batch mean directions are pairwise cosine-similar
at $0.988 \pm 0.005$ across eight different text batches.
\S\ref{sec:dc} characterizes that anchor.

\FloatBarrier

\subsection{The mean direction: a learned default-token distribution}
\label{sec:dc}

Projecting $\bar s$ through the model's LM head yields a sharp
distribution whose top-10 tokens are LLaMA's most frequent function
tokens (``,'', `` and'', `` the'', `` in'', `` to'', `` a''), with
Spearman correlation $+0.46$ against the training unigram. This is
not an artifact of averaging embeddings: the mean embedding decoded
the same way overlaps the $\bar s$ top-10 in $0/10$ tokens
(cosine $-0.62$). $\bar s$ is a direction training specifically
shaped.

$\bar s$ appears quickly --- the running mean of the state reaches
cosine $0.98$ with the final $\bar s$ by token 10, $0.996$ by token
100 --- and persists even when given a way out. Adding an explicit
learnable bias to the decoder (a 50M 1L~GPN, 8k steps) drops
$\|\bar s\|$ by only $6\%$ ($64.5 \to 60.3$); $\bar s$ remains the
top principal component and the learned bias decodes to the same
function-token top-10. Training keeps the default-token mechanism
in the state even when given an architectural alternative,
suggesting $\bar s$ carries structural roles beyond decoding (e.g.,
a magnitude anchor for RMSNorm) that a decode-only bias cannot
replace.

\FloatBarrier

\subsection{Two horizons from one memory rule}
\label{sec:horizon}

The state is rewritten every step; the matrix memory is a larger
decay-governed store written by the state and read back on demand.
We measure how long information survives in each.

\begin{figure}[!ht]
\centering
\includegraphics[width=\linewidth]{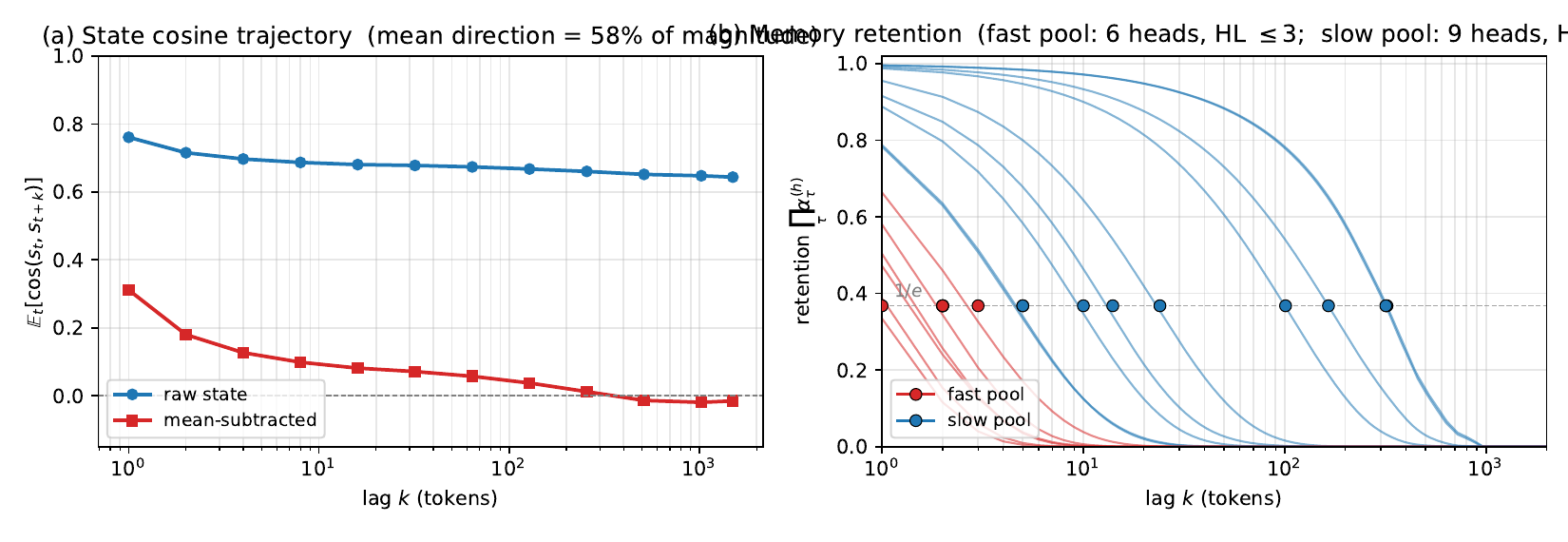}
\caption{Temporal horizons.
\textbf{(a)}~State cosine trajectory
$\mathbb{E}_t[\cos(s^p_t, s^p_{t+k})]$ vs.\ lag $k$ (log-$x$). The raw
state (blue) plateaus near $0.65$ up to $k=1500$ because the shared
mean direction dominates both vectors; after subtracting each
sequence's mean (red), the content-bearing component drops from
$0.31$ at $k{=}1$ to $\approx 0$ by $k \approx 256$.
\textbf{(b)}~Per-head retention $\prod_{\tau \le k} \alpha_\tau^{(h)}$
of the matrix memory $M$, one curve per head; dots mark each head's
$1/e$ half-life. Heads split into a fast pool (red, $1$--$3$ tokens)
and a slow pool (blue, up to $322$ tokens), roughly an order of
magnitude longer than the state's own content horizon.}
\label{fig:horizons}
\end{figure}

\paragraph{State horizon: tens of tokens.}
The raw cosine trajectory $c_k = \mathbb{E}_t[\cos(s^p_t, s^p_{t+k})]$
plateaus near $0.65$ out to $k = 1500$, which is $\bar s$ dominating
both endpoints, not genuine long-range memory. Once the per-sequence
mean is subtracted, the centered cosine falls from $0.31$ at $k=1$
to $\approx 0$ by $k = 256$ (Fig.~\ref{fig:horizons}a). The
content-bearing part of the state holds information for tens of
tokens.

\paragraph{Memory horizon: heads split into fast and slow pools.}
Each head $h$ applies a data-dependent decay
$\alpha_t^{(h)} \in (0,1)$ to $M^{(h)}_{t-1}$ before writing
(\S\ref{sec:architecture}); accumulating these decays along held-out
data gives per-head retention curves and a half-life $k_{1/e}$
(the lag at which $\prod_\tau \alpha_\tau^{(h)} = 1/e$). The 15 heads
split into two pools (Fig.~\ref{fig:horizons}b): a \emph{fast pool}
of 6 heads with half-lives of $1$--$3$ tokens (write-dominated
scratch), and a \emph{slow pool} of 9 heads with half-lives from $5$
to $322$ tokens. The slowest heads retain information roughly an
order of magnitude longer than the state itself.

All 15 heads are architecturally identical; nothing in the design
divides them into short- and long-retention pools. The division
emerges from training. We did not have to build a two-timescale
memory system: one associative memory with data-dependent decay is
enough, and the two timescales arise because training finds them
useful. Whether the slow pool is actually being used for
long-range prediction --- as opposed to merely being capable of it
--- requires an ablation we have not run.

\FloatBarrier

\section{Limitations}

This is a result about what a single shallow recurrent layer can
carry, not a production-ready system. Practical and scientific
caveats:

(1)~\textbf{Training cost.} GPN/GPN+M run sequentially in Python
across the time dimension; the in-loop matrix-memory update has no
parallel-along-sequence kernel, so per-step training cost is well
above the transformer and GDN baselines at matched parameter count.
This is a research artifact, not a deployable model.
(2)~\textbf{Scale.} One seed per configuration at 130M, $\sim 20$B
training tokens; no scaling sweeps.
(3)~\textbf{Descriptive analyses.} We do not measure the PPL impact
of zeroing $\bar s$ or ablating slow-pool memory heads, so the
emergent structures cannot be called causal.
(4)~\textbf{Baselines.} We do not run matched 1-layer GDN or Mamba
with the same measurements, so we cannot attribute the observed
structures to grounded prediction specifically.
(5)~\textbf{Depth cost on context-heavy tasks.} Averaged FineWeb PPL
hides what LAMBADA reveals --- limited per-token compute depth costs
accuracy on context-dependent predictions, even with memory.

\section{Conclusion}

GPN replaces the deep stack with a single recurrent layer revisited
at every step around one state vector and one shared matrix memory.
At 130M parameters, a 1-layer GPN+M reaches FineWeb PPL $18.06$,
$13\%/18\%$ above a 12-layer Transformer++ and a 10-layer GDN; a
2-layer variant brings the gap to $6\%/11\%$. The shallow model
does not match the deep baselines, and pays a measurable cost on
context-specific predictions like LAMBADA.

Conceptually, the state's prediction of the next step can be viewed
as part of the encoding of that step --- a single vector plays both
roles, with grounding completing the encoding. Because the working
state is a single vector, structures a layered architecture would
have to dig out of residual streams are here on the surface ---
small enough to see through, large enough to be empirically usable.

\bibliographystyle{plainnat}
\bibliography{refs}

\end{document}